\documentclass[12pt]{article}
\usepackage{geometry}
\usepackage{amsmath,amssymb}
\usepackage{graphicx}
\usepackage{booktabs}
\usepackage{caption}
\usepackage{subcaption}
\usepackage{cite}
\usepackage{url}
\usepackage{float}
\usepackage{bm}
\usepackage{times}   
\usepackage[colorlinks=true, linkcolor=blue, citecolor=blue, urlcolor=blue]{hyperref}

\begin{document}

\title{\textbf{Seismic full-waveform inversion based on a physics-driven generative adversarial network}}
\author{
Xinyi Zhang, Caiyun Liu\thanks{Corresponding author}, Jie Xiong, Qingfeng Yu \\
\textit{School of Electronic Information and Electrical Engineering, Yangtze University} \\
Jingzhou, Hubei 434023, P.R.China
}
\date{}
\maketitle

\begin{abstract}
\textbf{Objectives:} Full-waveform inversion (FWI) is a high-resolution geophysical imaging technique that reconstructs subsurface velocity models by iteratively minimizing the misfit between predicted and observed seismic data. However, under complex geological conditions, conventional FWI suffers from strong dependence on the initial model and tends to produce unstable results when the data are sparse or contaminated by noise.\\
\textbf{Methods:} To address these limitations, this paper proposes a physics-driven generative adversarial network–based full-waveform inversion method. The proposed approach integrates the data-driven capability of deep neural networks with the physical constraints imposed by the seismic wave equation, and employs adversarial training through a discriminator to enhance the stability and robustness of the inversion results.\\
\textbf{Results:} Experimental results on two representative benchmark geological models demonstrate that the proposed method can effectively recover complex velocity structures and achieves superior performance in terms of structural similarity (SSIM) and signal-to-noise ratio (SNR).\\
\textbf{Conclusions:} This method provides a promising solution for alleviating the initial-model dependence in full-waveform inversion and shows strong potential for practical applications.
\end{abstract}

\section{Introduction}
Seismic inversion is an essential tool for studying the Earth's internal structure and exploring natural resources. The velocity model plays a critical role in seismic data processing and interpretation, particularly in migration imaging, structural interpretation, and attribute analysis \cite{Yang2013}. Traditional velocity modeling methods mainly include velocity spectrum analysis \cite{Berkhout1997}, tomography \cite{Iyer1993}, and full-waveform inversion (FWI) \cite{Virieux2009}. Velocity spectrum analysis estimates layer velocities from reflection arrival times; it is simple but has low resolution. Tomography inverts travel-time residuals based on ray theory, improving accuracy but still limited in characterizing complex structures. FWI fully utilizes all wavefield information in seismic records and can invert high-resolution subsurface velocity structures, making it a focal point of current research.

Since Tarantola \cite{Tarantola1984} proposed FWI, it has shown significant advantages in high-resolution imaging due to its ability to iteratively update subsurface velocity structures using full wavefield information. However, traditional FWI usually relies on a good initial model and low-frequency data, easily falling into local minima and suffering from noise, cycle skipping, and other issues in practical applications, which limits its accuracy and stability \cite{Koren1991,Wang2015}. To address these nonlinear problems, Akcelik \cite{Akcelik2002} introduced a Gauss–Newton optimization method and applied it to a 2D acoustic wave synthetic model, improving inversion accuracy. Ren et al. \cite{Ren2013} systematically studied the Hessian operator in seismic inversion imaging. Wang and Dong \cite{WangYi2015} implemented multi-parameter acoustic FWI in VTI media using a truncated Newton method and obtained relatively accurate results. Liu et al. \cite{Liu2013} introduced a modified quasi-Newton condition in frequency-domain FWI, accelerating convergence. Miao \cite{Miao2015} applied the L-BFGS algorithm to time-domain FWI. Zhang et al. \cite{Zhang2013} implemented multi-scale FWI.

Furthermore, to overcome the local-minimum problem, simulated annealing \cite{Yang2010,Han2019}, genetic algorithms \cite{Kirkpatrick1983}, and particle swarm optimization \cite{Zhu2013} have been introduced into FWI to reduce dependence on the initial model. Pan et al. \cite{Pan2021} proposed a hybrid adaptive genetic algorithm (HAGA) to further improve search efficiency. Overall, although global optimization methods can partially overcome cycle skipping and reduce initial model dependence, they are computationally expensive and converge slowly, often requiring combination with local optimization strategies in practice.

In recent years, deep learning has been rapidly applied in seismic exploration, especially in velocity model building. Yang and Ma \cite{Yang2019} proposed a modified U-Net framework that can effectively predict velocity models from multiple seismic data, approximating the nonlinear relationship between seismic data and models. Wu and McMechan \cite{Wu2019} introduced a deep image prior (DIP)-based CNN-domain FWI method, updating network parameters by minimizing the error between simulated and observed data. Recurrent neural networks (RNNs) \cite{Mikolov2011} have been applied to seismic data reconstruction \cite{Yoon2020} and impedance inversion \cite{Alfarraj2019}, demonstrating their advantages in handling variable-length input sequences. However, most of these methods rely heavily on large amounts of labeled data and lack physical constraints, which may lead to inconsistencies between generated results and actual observations, limiting generalization. In addition, Raissi et al. \cite{Raissi2019} proposed physics-informed neural networks (PINNs), which efficiently solve nonlinear partial differential equations (PDEs) by combining mathematical models with data, and have been successfully applied to waveform inversion. Under an unsupervised learning framework, PINNs incorporate physical constraints to generate velocity models that both fit observed seismic data and satisfy physical laws, but they still depend on training data quality and network architecture design.

Based on the idea of physical constraints, Yang and Ma \cite{Yang2023} proposed physics-informed generative adversarial networks (FWIGAN), which apply adversarial constraints through a discriminator to traditional FWI results and achieved good inversion performance. However, in that method, the generator consists only of traditional FWI, and model updates rely mainly on physical gradient information; no learnable velocity model mapping is established, so under complex geological conditions, it is still inevitably constrained by the initial model. To address this problem, this paper improves the generator structure by introducing a deep neural network to parameterize the generator, enabling it to learn structural features of subsurface velocity models from observed seismic data and collaboratively optimize with the FWI iteration process under wave-equation constraints. On one hand, the deep neural network provides a structurally reasonable initial model for FWI, effectively reducing the sensitivity of inversion to the initial model. On the other hand, the physical inversion process corrects the network predictions under wave-equation constraints, ensuring the physical consistency of the generated velocity model. Meanwhile, FWI also provides additional training data for the deep neural network, giving the generator a "learnable" characteristic, and the deep neural network prediction and physical inversion process mutually promote each other, achieving joint optimization.

\section{Basic Principles}
\subsection{Traditional FWI}
Full-waveform inversion (FWI) \cite{Virieux2009} is a method for high-resolution reconstruction of subsurface velocity models by minimizing the difference between simulated and observed seismic data. This paper adopts the 2D constant-density acoustic wave equation (AWE) to describe wave propagation, expressed in the time domain as:
\begin{equation} \label{eq:wave}
\frac{\partial^2 u(\mathbf{x},t)}{\partial t^2} - v^2(\mathbf{x})\nabla^2 u(\mathbf{x},t) = f(t)\delta(\mathbf{x}-\mathbf{x}_s),
\end{equation}
where $u(\mathbf{x},t)$ represents the seismic wavefield, $v(\mathbf{x})$ is the velocity model, $f(t)$ is the source term, and $\delta(\mathbf{x}-\mathbf{x}_s)$ is the Dirac function indicating the source location. The forward modeled data are $d_{\text{syn}}(\mathbf{x}_r,t;\mathbf{x}_s)=\mathcal{R}u$, where $\mathcal{R}$ is a sampling operator, and the observed seismic data are denoted as $d_{\text{obs}}(\mathbf{x}_r,t;\mathbf{x}_s)$.

The least-squares misfit function is constructed as:
\begin{equation} \label{eq:fwi_loss}
E(v) = \frac{1}{2}\sum_{s=1}^{N_s}\sum_{r=1}^{N_r}\int_0^T \bigl[d_{\text{syn}}(\mathbf{x}_r,t;\mathbf{x}_s)-d_{\text{obs}}(\mathbf{x}_r,t;\mathbf{x}_s)\bigr]^2 dt,
\end{equation}
where $T$ is the maximum recording time, and $N_s$ and $N_r$ are the numbers of sources and receivers, respectively.

The goal of FWI is to iteratively update the velocity model $v$ by minimizing $E(v)$. In numerical implementation, the domain $\Omega$ of the velocity model $v$ is discretized into an $N_x\times N_z$ pixel grid, and the corresponding discretized model and data can be expressed as:
\begin{equation} \label{eq:discrete}
\mathbf{v}\in\mathbb{R}^{N_x N_z},\quad \mathbf{d}_{\text{syn}} = \mathcal{F}(\mathbf{v}),\quad \mathbf{d}_{\text{obs}}\in\mathbb{R}^{N_s N_r N_t},
\end{equation}
where $\mathbf{v}$ is the discretized velocity model, $\mathbf{d}_{\text{syn}}$ is the simulated seismic data, $\mathbf{d}_{\text{obs}}$ is the observed data, and $\mathcal{F}$ is the forward modeling operator constrained by the wave equation.

\subsection{GAN Architecture}
Generative adversarial networks (GANs) \cite{Goodfellow2014} are a typical class of generative deep learning models, consisting of two neural networks: a generator and a discriminator. The generator aims to generate samples that approximate the real data distribution from an input space, while the discriminator's task is to distinguish whether an input sample comes from real data or from the generator's fake data. The two networks are optimized through adversarial training, eventually enabling the generator to produce highly realistic samples. The overall loss consists of a generator loss and a discriminator loss, with opposing objectives. The overall optimization objective of GAN can be expressed as the following minimax problem:
\begin{equation} \label{eq:gan_minimax}
\min_G \max_D V(D,G) = \mathbb{E}_{x\sim p_{\text{data}}}[\log D(x)] + \mathbb{E}_{z\sim p_z}[\log(1-D(G(z)))],
\end{equation}
where $x$ represents real samples, $z$ is a noise vector, $p_{\text{data}}$ and $p_z$ are the distributions of real data and noise, respectively, $D(x)$ is the probability output by the discriminator, and $G(z)$ is the generator output.

Correspondingly, the loss functions of the discriminator and generator can be written as:
\begin{align}
\mathcal{L}_D &= -\mathbb{E}_{x\sim p_{\text{data}}}[\log D(x)] - \mathbb{E}_{z\sim p_z}[\log(1-D(G(z)))], \label{eq:gan_lossD}\\
\mathcal{L}_G &= -\mathbb{E}_{z\sim p_z}[\log D(G(z))]. \label{eq:gan_lossG}
\end{align}
The discriminator improves its ability to distinguish between real and generated samples by minimizing Eq.~\eqref{eq:gan_lossD}, while the generator continuously adjusts its parameters by minimizing Eq.~\eqref{eq:gan_lossG} so that its generated samples approach the real data distribution under the discriminator's judgment.

\section{Model Design}
\subsection{Overall Workflow}
\begin{figure}[htbp]
\centering
\includegraphics[width=1.0\textwidth]{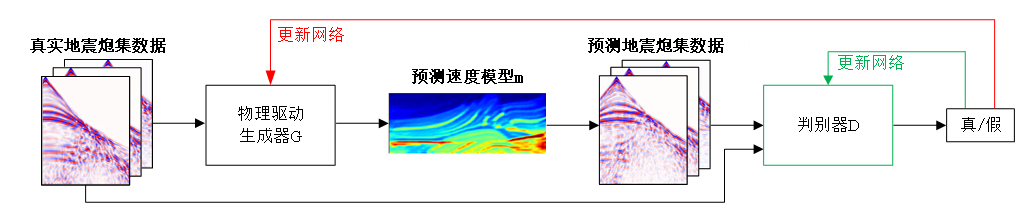}
\caption{Overall workflow}
\label{fig:workflow}
\end{figure}

The overall workflow of the proposed physics-driven generative adversarial network inversion method is shown in Fig.~\ref{fig:workflow}. The observed seismic shot gather data are first input into the physics-driven generator $G$, which produces a subsurface velocity model prediction that conforms to the physical laws of seismic wave propagation. Based on this predicted model, forward wave-equation modeling yields corresponding predicted seismic data, which are then input together with the observed seismic data into the discriminator $D$. The discriminator judges the authenticity of the input data, distinguishing between observed and predicted seismic data.

During adversarial training, the optimization goal of $G$ is to reduce the probability that its generated results are judged as "fake" by $D$, thereby guiding the generated velocity model to gradually approach the true subsurface structure. The goal of $D$ is to improve its ability to distinguish between observed and predicted seismic data, thereby forcing $G$ to further improve generation quality (i.e., inversion quality). Through alternating optimization between $G$ and $D$, the model continuously updates the generator parameters during iterations, achieving high-precision inversion of subsurface velocity models.

\subsection{Physics-Driven Generator}
\begin{figure}[htbp]
\centering
\includegraphics[width=1.0\textwidth]{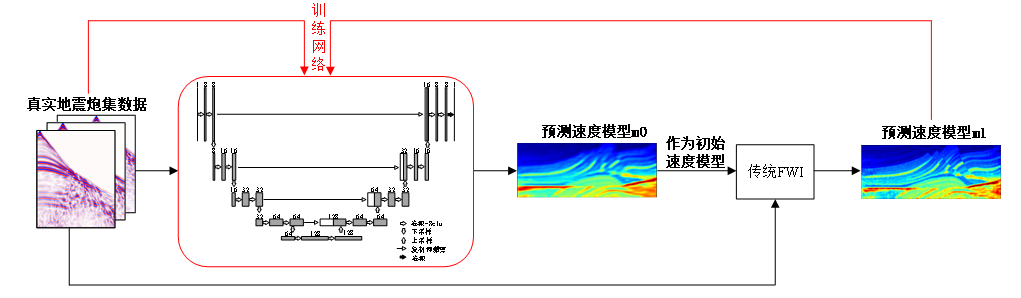}
\caption{Architecture of the physics-driven generator}
\label{fig:generator}
\end{figure}

The physics-driven generator $G$ consists of a U-Net structured deep neural network and an FWI solver, as shown in Fig.~\ref{fig:generator}. Unlike Yang and Ma \cite{Yang2023}, who used the traditional FWI iterative process as the generator, this paper parameterizes the generator by combining a deep neural network with the physical inversion process, constructing a learnable physics-driven generator structure.

The deep neural network in $G$ adopts the U-Net architecture, consisting of an encoder and a decoder. The encoder comprises four convolutional blocks, each containing two convolutional layers, ReLU activation, and max-pooling operations, extracting high-level semantic features through successive downsampling. A center layer is set at the end of the encoder to further enhance feature representation. The decoder consists of four upsampling blocks that progressively restore spatial resolution via transposed convolution or bilinear interpolation, and skip connections link feature maps from corresponding encoder layers to the decoder to fuse low-level detail information with high-level semantic information. Finally, a $1\times1$ convolutional layer outputs the predicted velocity model.

Before adversarial training, we first pre-train the U-Net in generator $G$ using only observed seismic data as input and the corresponding initial velocity model $v_{\text{init}}$ (obtained by linear interpolation or Gaussian smoothing) as the target, optimized with an L2 loss. This enables the network to learn generating a reasonable background velocity model from seismic data until the loss converges.

The adversarial training procedure of $G$ is as follows: First, observed seismic data are fed into the pre-trained U-Net, extracting multi-scale features from the seismic data through multi-layer nonlinear mappings, yielding an initial predicted subsurface velocity model $v_{\text{pre}}$. Then, $v_{\text{pre}}$ is used as the initial velocity model and input together with the observed seismic data into the traditional FWI process, obtaining a physically corrected velocity model $v_{\text{corr}}$.

\subsection{Discriminator}
\begin{figure}[htbp]
\centering
\includegraphics[width=0.5\textwidth]{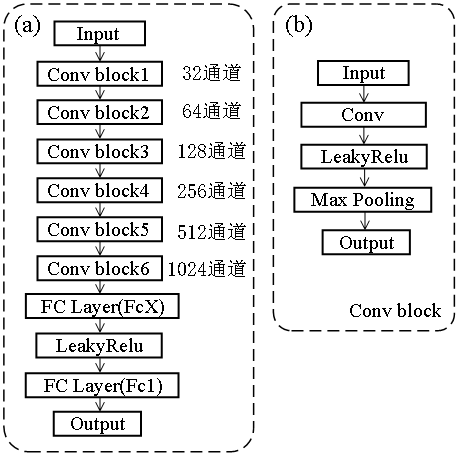}
\caption{Architecture of the discriminator: (a) Discriminator structure; (b) Convolutional block structure.}
\label{fig:discriminator}
\end{figure}

The discriminator $D$ adopts a convolutional neural network (CNN) structure, as shown in Fig.~\ref{fig:discriminator}, consisting of six convolutional blocks and two fully connected layers. Each convolutional block comprises a $3\times3$ convolutional layer, a Leaky ReLU activation with a negative slope of 0.1, and a $2\times2$ max-pooling layer that downsamples the feature maps to reduce spatial resolution. The first convolutional block has 32 channels, and the number of channels doubles in subsequent blocks. The input channel number is set according to the number of rays per batch of seismic data. The output of the last convolutional block is flattened into a vector and fed into fully connected layers, with neuron numbers set to 2000, 1000, and 1500 in different experiments (corresponding to FcX in Fig.~\ref{fig:discriminator}(a), where X indicates the number of neurons), processed by Leaky ReLU, and finally outputting a scalar representing the discrimination result. To ensure the effectiveness of the gradient penalty strategy, batch normalization is not used. The discriminator extracts features layer by layer and provides feedback, effectively pushing the generator to optimize its output and thus generate more realistic subsurface velocity models.

\section{Experiments}
\subsection{Experimental Setup}
The experiments in this paper were implemented using the PyTorch deep learning framework, and model training was conducted on a Linux system. The experimental platform was equipped with an NVIDIA Tesla P40 GPU (24GB memory) and 96GB of system RAM. The generator and discriminator were trained for 300 epochs with a batch size of 15 and a learning rate of 0.001. All networks were optimized using the Adam optimizer, with a training strategy of training the discriminator six times followed by training the generator once.

\subsection{Evaluation Metrics}
To quantitatively evaluate the inversion performance of the proposed method, the consistency between the predicted velocity model $\hat{\mathbf{v}}$ and the true velocity model $\mathbf{v}$ must be analyzed. The main evaluation metrics include structural similarity (SSIM) and signal-to-noise ratio (SNR), which comprehensively reflect the quality of the inverted model in terms of structural fidelity and noise suppression.

Structural similarity (SSIM) measures the similarity between the inverted model and the true model in terms of structure and texture features, defined as:
\begin{equation} \label{eq:ssim}
\text{SSIM}(\mathbf{v},\hat{\mathbf{v}}) = \frac{(2\mu_v\mu_{\hat{v}}+C_1)(2\sigma_{v\hat{v}}+C_2)}{(\mu_v^2+\mu_{\hat{v}}^2+C_1)(\sigma_v^2+\sigma_{\hat{v}}^2+C_2)},
\end{equation}
where $\mu_v$ and $\mu_{\hat{v}}$ are the local means of the two models, $\sigma_v$ and $\sigma_{\hat{v}}$ are their standard deviations, and the regularization constants are $C_1=(0.01L)^2$, $C_2=(0.03L)^2$ (with $L$ being the dynamic range of the grayscale values). A higher SSIM value indicates greater consistency in structural features between the inverted and true models.

To assess the impact of noise in the model, the signal-to-noise ratio (SNR) is introduced, defined as:
\begin{equation} \label{eq:snr}
\text{SNR} = 10\log_{10}\frac{\|\mathbf{v}\|_2^2}{\|\mathbf{v}-\hat{\mathbf{v}}\|_2^2}\quad\text{(dB)}.
\end{equation}
A larger SNR value indicates a higher proportion of effective signal in the inversion result, implying better model quality and stability.

\subsection{Marmousi Model Experiments}
\begin{figure}[htbp]
\centering
\includegraphics[width=1.0\textwidth]{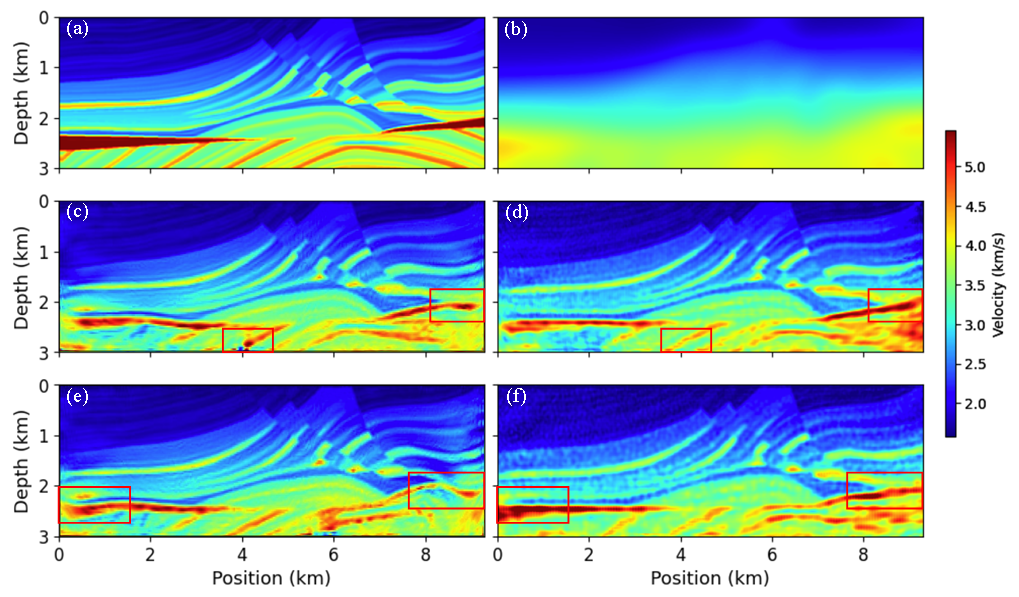}
\caption{Inversion results of the Marmousi model: (a) True velocity model; (b) Gaussian-smoothed initial velocity model; (c) Inversion result by Yang et al. using noise-free seismic data; (d) Inversion result by the proposed method using noise-free seismic data; (e) Inversion result by Yang et al. using noisy seismic data; (f) Inversion result by the proposed method using noisy seismic data.}
\label{fig:marmousi}
\end{figure}

The Marmousi model \cite{Versteeg1994} is one of the most representative acoustic velocity models in seismic forward and inverse modeling research. In this paper, a downsampled version of the Marmousi model was used as the research object, with a model size of $191\times 51$, a spatial grid spacing of 0.03 km, and a velocity range of 1472 m/s to 5772 m/s. The true velocity distribution is shown in Fig.~\ref{fig:marmousi}(a).

Under noise-free observation data conditions, the result of Yang and Ma \cite{Yang2023} (Fig.~\ref{fig:marmousi}(c)) obtained a relatively reasonable velocity model with a good initial model, but there were still noticeable artifacts and local anomalies in the deep region (indicated by the red box in Fig.~\ref{fig:marmousi}(c)), deviating somewhat from the true model. In contrast, the proposed method (Fig.~\ref{fig:marmousi}(d)) recovers the velocity structure more accurately and continuously; the inversion result not only clearly reconstructs the main geological interfaces but also preserves subsurface details well. When additive white Gaussian noise (AWGN) with an SNR of 10 dB was added to the observation data, the differences between the two methods became more pronounced. The method of Yang and Ma \cite{Yang2023} (Fig.~\ref{fig:marmousi}(e)) is sensitive to noise, and the inversion result exhibits local low-velocity anomalies and structural discontinuities (indicated by the red box in Fig.~\ref{fig:marmousi}(e)). In contrast, the proposed framework (Fig.~\ref{fig:marmousi}(f)) demonstrates stronger robustness and denoising capability under noise interference, yielding a smoother and more reasonable velocity distribution. The SSIM and SNR results between the true velocity model and the inverted velocity models are shown in Table~\ref{tab:marmousi}. The proposed method outperforms that of Yang and Ma in all metrics, demonstrating its effectiveness.

\begin{table}[htbp]
\centering
\caption{Statistical performance of Marmousi model inversion}
\label{tab:marmousi}
\begin{tabular}{lccc}
\toprule
Noise condition & Metric & Proposed method & Yang et al.'s method \\
\midrule
Noise-free & SSIM & \textbf{0.7092} & 0.6802 \\
           & SNR  & \textbf{19.37}  & 17.29 \\
Noisy      & SSIM & \textbf{0.6652} & 0.6301 \\
           & SNR  & \textbf{21.42}  & 16.71 \\
\bottomrule
\end{tabular}
\end{table}

\subsection{Overthrust Model Experiments}
\begin{figure}[htbp]
\centering
\includegraphics[width=1.0\textwidth]{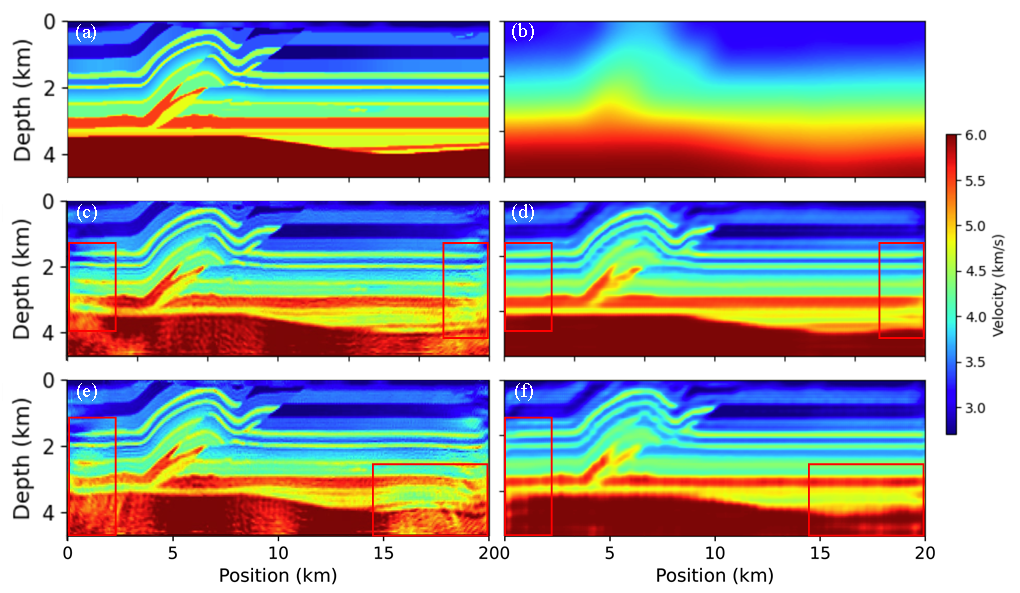}
\caption{Inversion results of the Overthrust model: (a) True velocity model; (b) Gaussian-smoothed initial velocity model; (c) Inversion result by Yang et al. using noise-free seismic data; (d) Inversion result by the proposed method using noise-free seismic data; (e) Inversion result by Yang et al. using noisy seismic data; (f) Inversion result by the proposed method using noisy seismic data.}
\label{fig:overthrust}
\end{figure}

The Overthrust model \cite{Aminzadeh1994} is a typical 3D geological velocity model. In this paper, the central slice of this 3D model was selected as the 2D research object. After downsampling, the model size is $251\times 81$, with a spatial step of 0.05 km and a velocity range of 2360 m/s to 6000 m/s. The true velocity distribution is shown in Fig.~\ref{fig:overthrust}(a).

Figure~\ref{fig:overthrust} displays the inversion results of the Overthrust model. This model contains distinct overthrust structures and a deep high-velocity basement, posing high demands on the structural resolution capability of the inversion method. The comparison results show that, in both noise-free and noisy cases, the imaging quality of Yang and Ma's method \cite{Yang2023} is limited in deep and laterally varying regions (indicated by red boxes in Fig.~\ref{fig:overthrust}(c) and (e)), with insufficiently clear local structural shapes. In contrast, the proposed method recovers a more continuous velocity distribution beneath the overthrust structures and in the deep high-velocity zone, with more complete representation of structural interfaces, indicating better imaging stability under complex structural conditions. The SSIM and SNR evaluations of the inverted models are given in Table~\ref{tab:overthrust}.

\begin{table}[htbp]
\centering
\caption{Statistical performance of Overthrust model inversion}
\label{tab:overthrust}
\begin{tabular}{lccc}
\toprule
Noise condition & Metric & Proposed method & Yang et al.'s method \\
\midrule
Noise-free & SSIM & \textbf{0.7842} & 0.7270 \\
           & SNR  & \textbf{28.57}  & 25.83 \\
Noisy      & SSIM & \textbf{0.7286} & 0.6889 \\
           & SNR  & \textbf{27.52}  & 23.88 \\
\bottomrule
\end{tabular}
\end{table}

\subsection{Inversion Experiments with a Linear Initial Velocity Model}
\begin{figure}[htbp]
\centering
\includegraphics[width=1.0\textwidth]{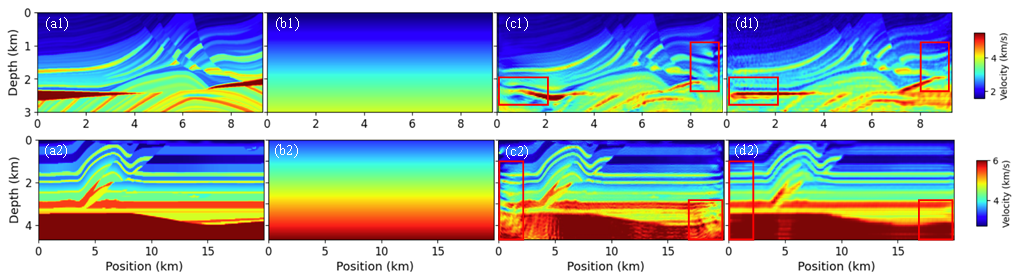}
\caption{Inversion results using a linear initial model: (a1–d1) True velocity model, linear initial velocity model, inversion result by Yang et al., and inversion result by the proposed method for the Marmousi model; (a2–d2) True velocity model, linear initial velocity model, inversion result by Yang et al., and inversion result by the proposed method for the Overthrust model.}
\label{fig:linear_init}
\end{figure}

Figure~\ref{fig:linear_init} shows the inversion results of different methods on the two complex models (Marmousi and Overthrust) when using a linear initial velocity model. Because the initial model significantly deviates from the true model, the method of Yang and Ma \cite{Yang2023} has limitations in depicting structural details in fault-dense zones and deep regions (indicated by red boxes in Fig.~\ref{fig:linear_init}(c1) and (c2)), with relatively weak continuity of local velocity structures. In contrast, the proposed method, under the same initial conditions, gradually converges to a reasonable velocity distribution and obtains clearer imaging results at multiple key structural locations, demonstrating lower dependence on the initial model and better adaptability in complex inversion problems. The SSIM and SNR evaluations of the inverted models are given in Table~\ref{tab:linear}.

\begin{table}[htbp]
\centering
\caption{Statistical performance of inversion using a linear initial model}
\label{tab:linear}
\begin{tabular}{lccc}
\toprule
Model & Metric & Proposed method & Yang et al.'s method \\
\midrule
Marmousi   & SSIM & \textbf{0.6582} & 0.6550 \\
           & SNR  & \textbf{18.21}  & 15.35 \\
Overthrust & SSIM & \textbf{0.7607} & 0.7377 \\
           & SNR  & \textbf{28.50}  & 23.92 \\
\bottomrule
\end{tabular}
\end{table}

\section{Conclusions}
This paper proposes a physics-driven generative adversarial network for seismic full-waveform inversion. By combining traditional FWI with generative adversarial networks, the method achieves high-precision velocity model inversion. The approach effectively alleviates the dependence of deep learning methods on large amounts of labeled data and reduces the sensitivity of traditional FWI to the initial model. Numerical experiments on two typical complex models, Marmousi and Overthrust, demonstrate that the proposed method can stably recover complex geological structures under both noise-free and noisy conditions, exhibiting better imaging quality and robustness in fault-dense zones, deep high-velocity regions, and areas with strong lateral velocity variations. Comprehensive qualitative comparisons and quantitative evaluations indicate that the proposed method outperforms the compared method in inversion accuracy, stability, and noise resistance, showing great potential for high-precision velocity inversion under complex geological conditions.

\section*{Acknowledgements}
This paper is the result of the National Natural Science Foundation of China funded project (No. 62273060).

\end{document}